\documentclass{llncs}

\usepackage{mathptmx}
\usepackage{multirow}
\usepackage{amsmath,bm}
\usepackage{graphicx}
\usepackage{subfigure}
\usepackage{array}
\usepackage{makeidx}  
\usepackage{expl3}
\usepackage{url}
\usepackage{multirow}
\usepackage{booktabs}

\usepackage{newtxtext,newtxmath}

\ExplSyntaxOn
\newcommand\latinabbrev[1]{
  \peek_meaning:NTF . {
    #1\@}%
  { \peek_catcode:NTF a {
      #1.\@ }%
    {#1.\@}}}
\ExplSyntaxOff

\begin{document}

%

%
\mainmatter              
\title{Medical Image Synthesis for Data Augmentation and Anonymization using Generative Adversarial Networks}
%
%
\author{Hoo-Chang Shin\inst{1} \and Neil A Tenenholtz\inst{2} \and Jameson K Rogers\inst{2} \and Christopher G Schwarz\inst{3} \and Matthew L Senjem\inst{3} \and Jeffrey L Gunter\inst{3} \and Katherine Andriole\inst{2} \and Mark Michalski\inst{2}}
%
%
\tocauthor{Ivar Ekeland, Roger Temam, Jeffrey Dean, David Grove, Craig Chambers, Kim B. Bruce, and Elisa Bertino}
\institute{NVIDIA Corporation\\
\and
MGH \& BWH Center for Clinical Data Science, Boston, MA, USA\\
\and
Mayo Clinic, Rochester, MN, USA\\
\footnotesize{\email{hshin@nvidia.com, jrogers24@partners.org, gunter.jeffrey@mayo.edu}}
}

\maketitle              

\begin{abstract}
Data diversity is critical to success when training deep learning models.
Medical imaging data sets are often imbalanced as pathologic findings are generally rare, which introduces significant challenges when training deep learning models.
In this work, we propose a method to generate synthetic abnormal MRI images with brain tumors by training a generative adversarial network using two publicly available data sets of brain MRI.
We demonstrate two unique benefits that the synthetic images provide.
First, we illustrate improved performance on tumor segmentation by leveraging the synthetic images as a form of data augmentation.
Second, we demonstrate the value of generative models as an anonymization tool, achieving comparable tumor segmentation results when trained on the synthetic data versus when trained on real subject data.
Together, these results offer a potential solution to two of the largest challenges facing machine learning in medical imaging, namely the small incidence of pathological findings, and the restrictions around sharing of patient data.
\end{abstract}
\section{Introduction}

It is widely known that sufficient data volume is necessary for training a successful machine learning algorithm \cite{domingos2012few} for medical image analysis.
Data with high class imbalance or of insufficient variability \cite{shin2016deep} leads to poor classification performance.
This often proves to be problematic in the field of medical imaging where abnormal findings are by definition uncommon.
Moreover, in the case of image segmentation tasks, the time required to manually annotate volumetric data only exacerbates this disparity; manually segmenting an abnormality in three dimensions can require upwards of fifteen minutes per study making it impractical in a busy radiology practice.
The result is a paucity of annotated data and considerable challenges when attempting to train an accurate algorithm.
While traditional data augmentation techniques (e.g., crops, translation, rotation) can mitigate some of these issues, they fundamentally produce highly correlated image training data.

In this paper we demonstrate one potential solution to this problem by generating synthetic images using a generative adversarial network (GAN) \cite{goodfellow2014generative}, which provides an additional form of data augmentation and also serves as a effective method of data anonymization.
Multi-parametric magnetic resonance images (MRIs) of abnormal brains (with tumor) are generated from segmentation masks of brain anatomy and tumor.
This offers an automatable, low-cost source of diverse data that can be used to supplement the training set.
For example, we can alter the tumor's size, change its location, or place a tumor in an otherwise healthy brain, to systematically have the image and the corresponding annotation.
Furthermore, GAN trained on a hospital data to generate synthetic images can be used to share the data outside of the institution, to be used as an anonymization tool.

Medical image simulation and synthesis have been studied for a while and are increasingly getting traction in medical imaging community \cite{8305584}.
It is partly due to the exponential growth in data availability, and partly due to the availability of better machine learning models and supporting systems.
Twelve recent research on medical image synthesis and simulation were presented in the special issue of Simulation and Synthesis in Medical Imaging \cite{8305584}.

This work falls into the synthesis category, and most related works are those of Chartsias \textit{et al} \cite{8071026} and Costa \textit{et al} \cite{8055572}.
We use the publicly available data set (ADNI and BRATS) to demonstrate multi-parametric MRI image synthesis and Chartsias \textit{et al} \cite{8071026} use BRATS and ISLES (Ischemic Stroke Lesion Segmentation (ISLES) 2015 challenge) data set.
Nonetheless, evaluation criteria for synthetic images were demonstrated on MSE, SSIM, and PSNR, but not directly on diagnostic quality.
Costa \textit{et al} \cite{8055572} used GAN to generate synthetic retinal images with labels, but the ability to represent more diverse pathological pattern was limited compared to this work.
Also, both previous works were demonstrated on 2D images or slices/views of 3D images, whereas in this work we directly process 3D input/output.
The input/output dimension is 4D when it is multi-parametric (T1/T2/T1c/Flair).
We believe processing data as 3D/4D in nature better reflects the reality of data and their associated problems.

Reflecting the general trend of the machine learning community, the use of GANs in medical imaging has increased dramatically in the last year.
GANs have been used to generate a motion model from a single preoperative MRI \cite{hu2017intraoperative}, upsample a low-resolution fundus image \cite{mahapatra2017image}, create a synthetic head CT from a brain MRI \cite{nie2017medical}, and synthesizing T2-weight MRI from T1-weighted ones (and vice-versa) \cite{dar2018image}.
Segmentation using GANs was demonstrated in \cite{zhang2017deep,yang2017automatic}.
Finally, Frid-Adar \textit{et al}. leveraged a GAN for data augmentation, in the context of liver lesion classification \cite{frid2018synthetic}.
To the best of our knowledge, there is no existing literature on the generation of synthetic medical images as form of anonymization and data augmentation for tumor segmentation tasks.

\section{Data}
\label{sec:data}
\subsection{Dataset}
We use two publicly available data set of brain MRI:
\\\\
\textbf{Alzheimer's Disease Neuroimaging Initiative (ADNI) data set}\\
The ADNI was launched in 2003 as a public-private partnership, led by principal investigator Michael W. Weiner, MD.
The primary goal of ADNI has been to test whether serial magnetic resonance imaging (MRI), positron emission tomography (PET), other biological markers, and clinical and neuropsychological assessment can be combined to measure the progression of mild cognitive impairment (MCI) and early Alzheimer's disease (AD).
For up-to-date information on the ADNI study, see \url{www.adni-info.org}.
We follow the approach of \cite{schwarz2016large} that is shown to be effective for segmenting the brain atlas of ADNI data.
The atlas of white matter, gray matter, and cerebrospinal fluid (CSF) in the ADNI T1-weighted images are generated using the SPM12 \cite{ashburner2005unified} segmentation and the ANTs SyN \cite{tustison2014large} non-linear registration algorithms.
In total, there are 3,416 pairs of T1-weighted MRI and their corresponding segmented tissue class images.
\\
\\
\textbf{Multimodal Brain Tumor Image Segmentation Benchmark (BRATS) data set}\\
BRATS utilizes multi-institutional pre-operative MRIs and focuses on the segmentation of intrinsically heterogeneous (in appearance, shape, and histology) brain tumors, namely gliomas \cite{menze2015multimodal}.
Each patient's MRI image set includes a variety of series including T1-weighted, T2-weighted, contrast-enhanced T1, and FLAIR, along with a ground-truth voxel-wise annotation of edema, enhancing tumor, and non-enhancing tumor.
For more details about the BRATS data set, see \url{braintumorsegmentation.org}.
While the BRATS challenge is held annually, we used the BRATS 2015 training data set which is publicly available.

\subsection{Dataset Split and Pre-Processing}

As a pre-processing step, we perform skull-stripping \cite{iglesias2011robust} on the ADNI data set as skulls are not present in the BRATS data set.
The BRATS 2015 training set provides 264 studies, of which we used the first 80\% as a training set, and the remaining 20\% as a test set to assess final algorithm performance.
Hyper-parameter optimization was performed within the training set and the test set was evaluated only once for each algorithm and settings assessed.
Our GAN operates in 3D, and due to memory and compute constraints, training images were cropped axially to include the central 108 slices, discarding those above and below this central region, then resampled to $128\times 128\times 54$ for model training and inference.
For a fair evaluation of the segmentation performance to the BRATS challenge we used the original images with a resolution of $256\times 256\times 108$ for evaluation and comparison.
However, it is possible that very small tumor may get lost by the downsampling, thus affecting the final segmentation performance.

\begin{figure}[t!]
\centering
  \includegraphics[width=1\linewidth]{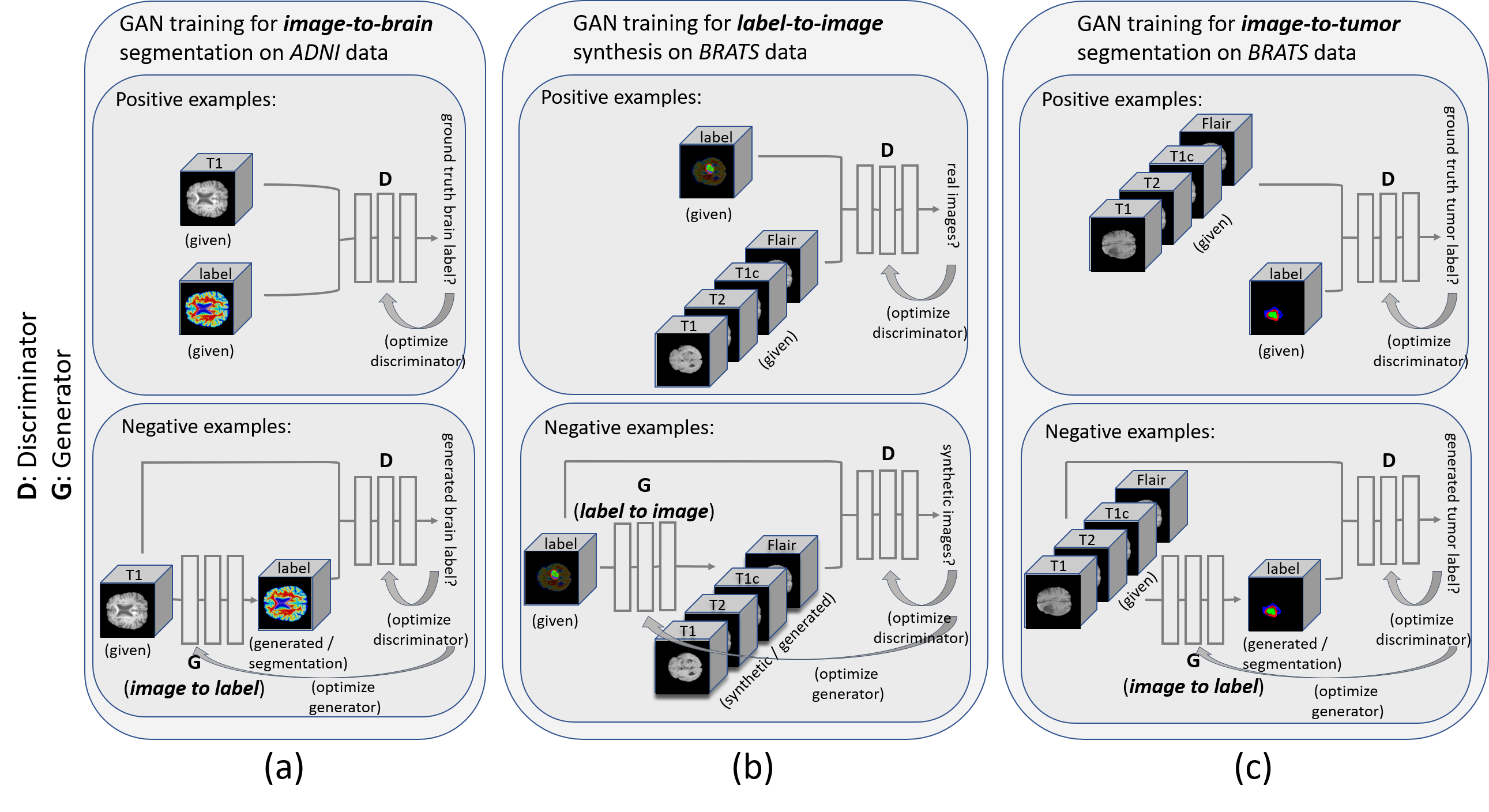}\\
  \caption{Illustration of training GAN for (a) MRI-to-brain segmentation; (b) label-to-MRI synthesis; (c) MRI-to-tumor segmentation.}
\label{fig:GAN-workflows}
\vspace{-5pt}
\end{figure}

\section{Methods}
\label{sec:method}

The image-to-image translation conditional GAN (\texttt{pix2pix}) model introduced in \cite{Isola_2017_CVPR} is adopted to translate label-to-MRI (synthetic image generation) and MRI-to-label (image segmentation).
For brain segmentation, the generator \textit{G} is given a T1-weighted image of ADNI as input and is trained to produce a brain mask with white matter, grey matter and CSF.
The discriminator \textit{D} on the other hand, is trained to distinguish ``real'' labels versus synthetically generated ``fake'' labels.
During the procedure (depicted in Figure~\ref{fig:GAN-workflows} (a)) the generator \textit{G} learns to segment brain labels from a T1-weighted MRI input.
Since we did not have an appropriate off-the-shelf segmentation method available for brain anatomy in the BRATS data set, and the ADNI data set does not contain tumor information, we first train the \texttt{pix2pix} model to segment normal brain anatomy from the T1-weighted images of the ADNI data set.
We then use this model to perform inference on the T1 series of the BRATS data set.
The segmentation of neural anatomy, in combination with tumor segmentations provided by the BRATS data set, provide a complete segmentation of the brain with tumor.

The synthetic image generation is trained by reversing the inputs to the generator and training the discriminator to perform the inverse task (i.e., ``is this imaging data acquired from a scanner or synthetically generated?'' as opposed to ``is this segmentation the ground-truth annotation or synthetically generated?'' -- Figure~\ref{fig:GAN-workflows} (b)).
We generate synthetic abnormal brain MRI from the labels and introduce variability by adjusting those labels (e.g., changing tumor size, moving the tumor's location, or placing tumor on a otherwise tumor-free brain label).
Then GAN segmentation module is used once again, to segment tumor from the BRATS data set (input: multi-parametric MRI; output: tumor label).
We compare the segmentation performance \textit{1)} with and without additional synthetic data, \textit{2)} using only the synthetic data and fine-tuning the model on 10\% of the real data; and compare their performance of GAN to a top-performing algorithm\footnote{\url{https://github.com/taigw/brats17}} \cite{wang2017automatic}from the BRATS 2017 challenge.

\begin{figure}[t!]
\centering
  \includegraphics[width=1\linewidth]{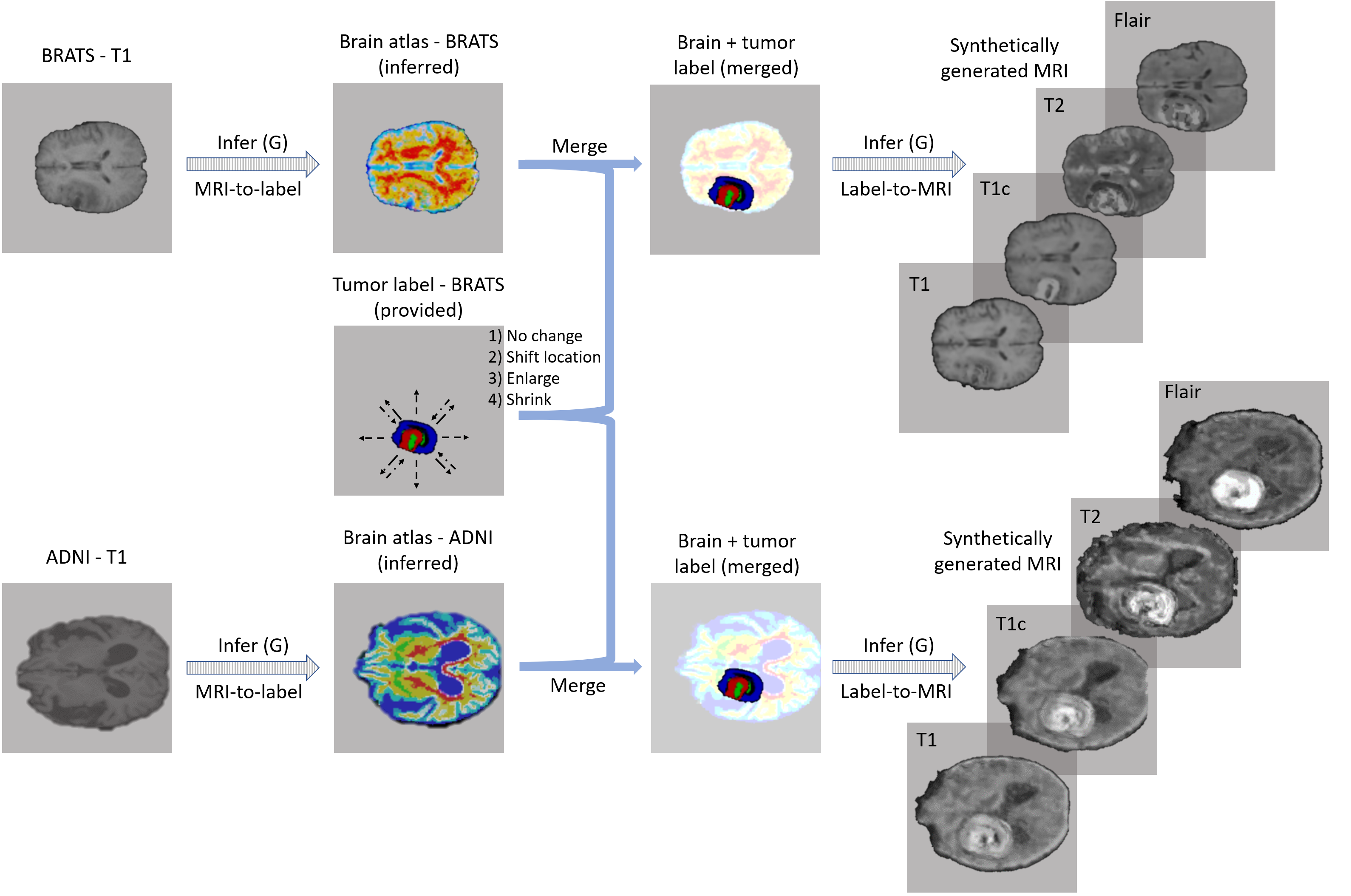}\\
  \caption{Workflow of getting synthetic images with variation. On BRATS data set, MRI-to-label image translation GAN is applied to T1-weighted images to get brain atlas. It is then merged with the tumor label given in the BRATS data set, possibly with alterations (shift tumor location; enlarge; shrink). The merged labels (with possibly alterations) are then used as an input to label-to-MRI GAN, to generate synthetic multi-parametric MRI with brain tumor.}
\label{fig:tm_manipulate_workflow}
\vspace{-5pt}
\end{figure}

\subsection{Data Augmentation with Synthetic Images}

\begin{figure}[t!]
\centering
  \includegraphics[width=1\linewidth]{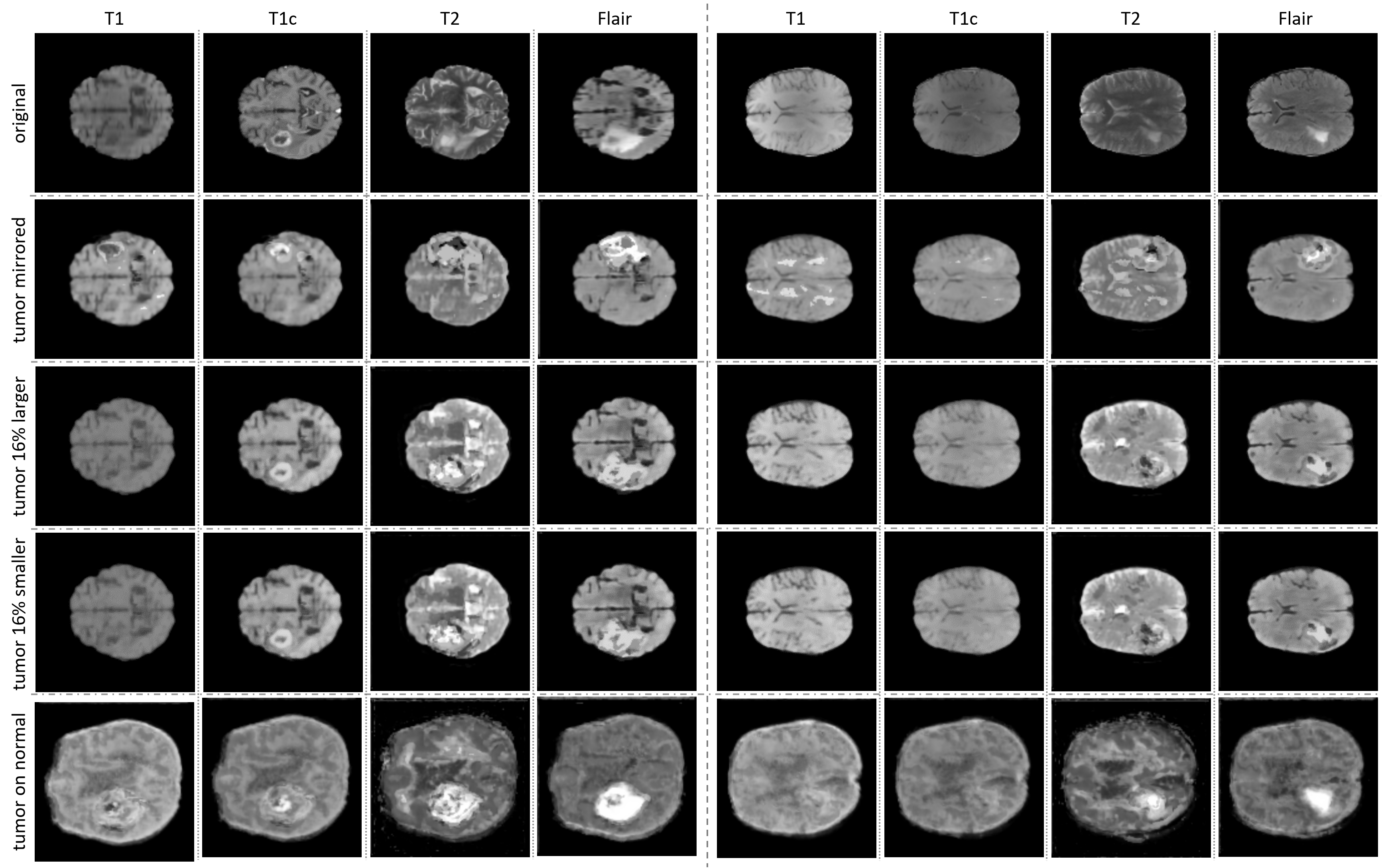}\\
  \caption{Examples of generated images. The first row depicts the original (``real'') images on which the synthetic tumors were based. Generated images without adjustment of the segmentation label are shown in the second row. Examples of generated images with various adjustments to the tumor segmentation label are shown in the third through fifth rows. The last row depicts examples of synthetic images where a tumor label is placed on a tumor-free brain label from the ADNI data set.}
\label{fig:tm_manipulate_examples}
\vspace{-5pt}
\end{figure}

The GAN trained to generate synthetic images from labels allows for the generation of arbitrary multi-series abnormal brain MRIs.
Since we have the brain anatomy label and tumor label separately, we can alter either the tumor label or the brain label to get synthetic images with the characteristics we desire.
For instance, we can alter the tumor characteristics such as size, location of the existing brain and tumor label set, or place tumor label on an otherwise tumor-free brain label.
Examples of this are shown in Figure~\ref{fig:tm_manipulate_examples}.

The effect of the brain segmentation algorithm's performance has not been evaluated in this study.

Since the GAN was first trained on 3,416 pairs of T1-weighted (T1) images from the ADNI data set, generated T1 images are of the high quality, and, qualitatively difficult to distinguish from their original counterparts.
BRATS data was used to train the generation of non-T1-weighted image series.
Contrast-enhanced T1-weighted images use the same image acquisition scheme as T1-weighted images.
Consequently, the synthesized contrast-enhanced T1 images appear reasonably realistic, although higher contrast along the tumor boundary is observed in some of the generated images.
T2-weighted (T2) and FLAIR image acquisitions are fundamentally different from the T1-weighted images, resulting in synthetic images that are less challenging to distinguish from scanner-acquired images.
However, given a sufficiently large training set on all these modalities, this early evidence suggests that the generation of realistic synthetic images on all the modalities may be possible.

Other than increasing the image resolution and getting more data especially for the sequences other than T1-weighted images, there are still a few important avenues to explore to improve the overall image quality.
For instance, more attention likely needs to be paid for the tumor boundaries so it does not look superimposed and discrete when synthetic tumor is placed.
Also, performance of brain segmentation algorithm and its ability to generalize across different data sets needs to be examined to obtain higher quality synthetic images combining data sets from different patient population.

\begin{figure}[t!]
\centering
  \includegraphics[width=.95\linewidth]{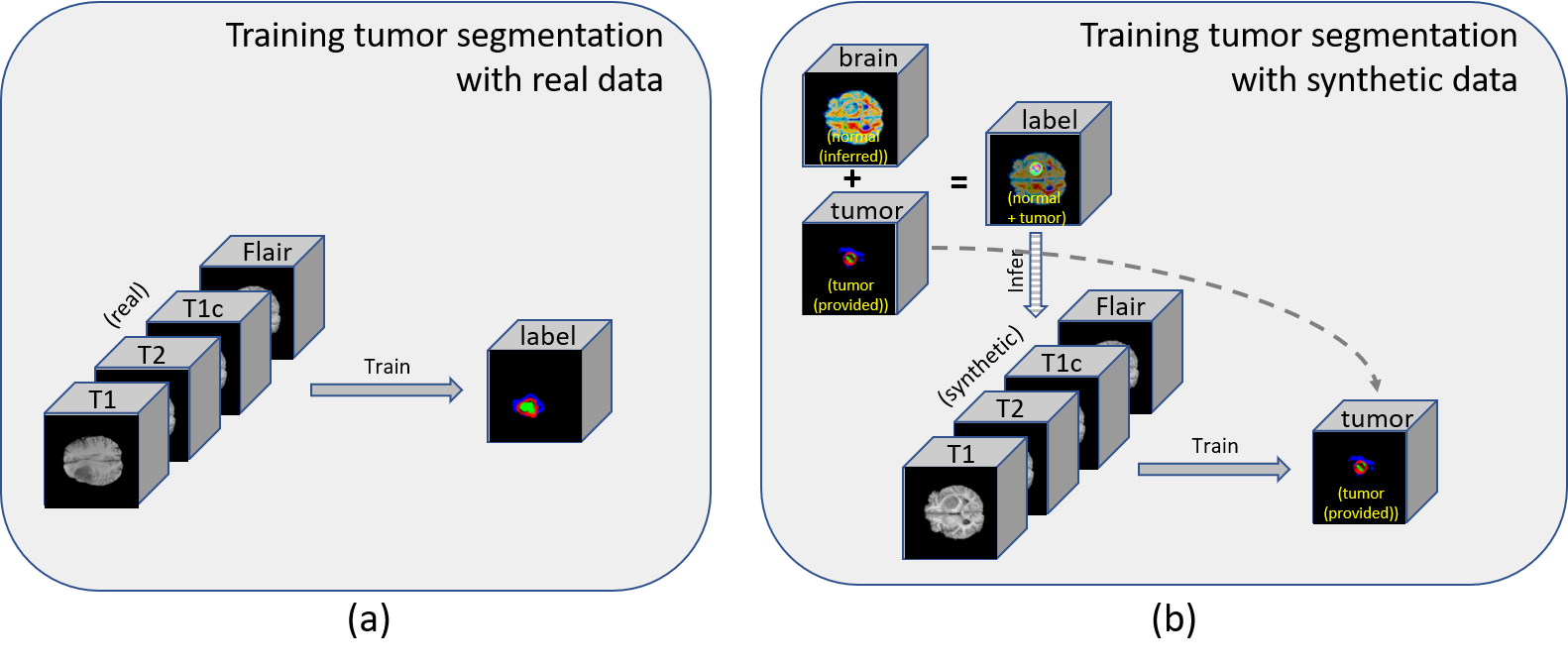}\\
  \caption{Training GAN for tumor segmentation with (a) real; and (b) synthetic image-label pairs. Synthetic data generation can increase the training data set with desired characteristics (e.g., tumor size, location, etc.) without the need of labor-intensive manual annotation.}
  \label{fig:real_synthetic_training}
\vspace{-5pt}
\end{figure}

The augmentation using synthetic images can be used in addition to the usual data augmentation methods such as random cropping, rotation, translation, or elastic deformation \cite{milletari2016v}.
Moreover, we have more control over the augmented images using the GAN-based synthetic image generation approach, that we have more input-option (i.e., label) to perturb the given image than the usual data augmentation techniques.
The usual data augmentation methods rely mostly on random processes and operates on the whole image level than specific to a location, such as tumor.
Additionally, since we generate image from the corresponding label, we get more images for training without needing to go through the labor-intensive manual annotation process.
Figure~\ref{fig:real_synthetic_training} shows the process of training GAN with real and synthetic image and label pairs.

\subsection{Generating Anonymized Synthetic Images with Variation}

Protection of personal health information (PHI) is a critical aspect of working with patient data.
Often times concern over dissemination of patient data restricts the data availability to the research community, hindering development of the field.
While removing all DICOM metadata and skull-stripping will often eliminate nearly all identifiable information, demonstrably proving this to a hospital's data sharing committee is near impossible.
Simply de-identifying the data is insufficient.
Furthermore, models themselves are subject to caution when derived from sensitive patient data.
It has been shown \cite{carlini2018secret} that private data can be extracted from a trained model.

Development of a GAN that generates synthetic, but realistic, data may address these challenges.
The first two rows of Figure~\ref{fig:tm_manipulate_examples} illustrate how, even with the same segmentation mask, notable variations can be observed between the generated and original studies.
This indicates that the GAN produces images that do not reflect the underlying patients as individuals, but rather draws individuals from the population in aggregate.
It generates new data that cannot be attributed to a single patient but rather an instantiation of the training population conditioned upon the provided segmentation.

\begin{table}[t]
\vspace{-5pt}
\renewcommand{\arraystretch}{1.3}
\renewcommand{\multirowsetup}{\centering}
\setlength{\belowrulesep}{0pt}
\setlength{\aboverulesep}{0pt}
\caption{Dice score evaluation (mean / standard deviation) of GAN-based segmentation algorithm and BRATS'17 top-performing algorithm \cite{wang2017automatic}, trained on ``real'' data only; real + synthetic data; and training on synthetic data only and fine-tuning the model on 10\% of the real data. GAN-based models were trained both with (with aug) and without (no aug) including the usual data augmentation techniques (crop, rotation, translation, and elastic deformation) during training. All models were trained for 200 epochs to convergence.}
\label{tab:results_gan_sota}
\begin{center}
\begin{tabular}{|c|||c|c||c|c|}
\hline
\multirow{2}{*}{Method} & \multirow{2}{*}{Real} & \multirow{2}{*}{Real + Synthetic} & \multirow{2}{*}{Synthetic only} & Synthetic only, \\
       &      &                  &                & fine-tune on 10\% real \\ \hline
\cmidrule{1-5}
GAN-based (no aug) & 0.64/0.14 & 0.80/0.07 & 0.25/0.14 & 0.80/0.18 \\
\hline
GAN-based (with aug) & 0.81/0.13 & 0.82/0.08 & 0.44/0.16 & 0.81/0.09 \\
\hline
Wang \textit{et al}. \cite{wang2017automatic} & 0.85/0.15 & 0.86/0.09 & 0.66/0.13 & 0.84/0.15 \\
\hline
\end{tabular}
\end{center}
\vspace{-5pt}
\end{table}

\section{Experiments and Results}

\subsection{Data Augmentation using Synthetic Data}

Dice score evaluation of the whole tumor segmentation produced by the GAN-based model and the model of Wang \textit{et al}. \cite{wang2017automatic} (trained on real and synthetic data) are shown in Table~\ref{tab:results_gan_sota}.
The segmentation models are trained on 80\% of the BRATS'15 training data only, and the training data supplemented with synthetic data.
Dice scores are evaluated on the 20\% held-out set from the BRATS'15 training data.
All models are trained for 200 epochs on NVIDIA DGX systems.

A much improved performance with the addition of synthetic data is observed without usual data augmentation (crop, rotation, elastic deformation; GAN-based (no-aug)).
However, a small increase in performance is observed when added with usual data augmentation (GAN-based (no-aug)), and it applies also to the model of Wang \textit{et al}. \cite{wang2017automatic} that incorporates usual data augmentation techniques.

Wang \textit{et al.} model operates in full resolution (256x256) combining three 2D models for each axial/coronal/sagittal view, whereas our model and generator operates in half the resolution (128x128x54) due to GPU memory limit.
We up-sampled the GAN-generated images twice the generated resolution for a fair comparison with BRATS challenge, however it is possible that very small tumor may get lost during the down-/up- sampling.
A better performance may be observed using the GAN-based model with an availability of GPU with more memory.
Also, we believe that the generated synthetic images having half the resolution, coupled with the lack of the image sequences for training other than T1-weighted ones possibly led to the relatively small increase in segmentation performance compared to using the usual data augmentation techniques.
We carefully hypothesize that with more T2/Flair images being available, better image quality will be observed for these sequences and so better performance for more models and tumor types.


\subsection{Training on Anonymized Synthetic Data}

We also evaluated the performance of the GAN-based segmentation on synthetic data only, in amounts greater than or equal to the amount of real data but without including any of the original data.
The dice score evaluations are shown in Table~\ref{tab:results_gan_sota}.
Sub-optimal performance is achieved for both our GAN-based and the model of Wang \textit{et al}. \cite{wang2017automatic} when training on an amount of synthetic data equal to the original 80\% training set.
However, higher performance, comparable to training on real data, is achieved when training the two models using more than five times as much synthetic data (only), and fine-tuning using a 10\% random selection of the ``real'' training data.
In this case, the synthetic data provides a form of pre-training, allowing for much less ``real'' data to be used to achieve a comparable level of performance.

\section{Conclusion}
\label{sec:conclusion}

In this paper, we propose a generative algorithm to produce synthetic abnormal brain tumor multi-parametric MRI images from their corresponding segmentation masks using an image-to-image translation GAN.
High levels of variation can be introduced when generating such synthetic images by altering the input label map.
This results in improvements in segmentation performance across multiple algorithms.
Furthermore, these same algorithms can be trained on completely anonymized data sets allowing for sharing of training data.
When combined with smaller, institution-specific data sets, modestly sized organizations are provided the opportunity to train successful deep learning models.

%
%
\bibliographystyle{plain}
\bibliography{main}

\clearpage

\begin{figure}[t!]
\centering
  \includegraphics[width=1\linewidth]{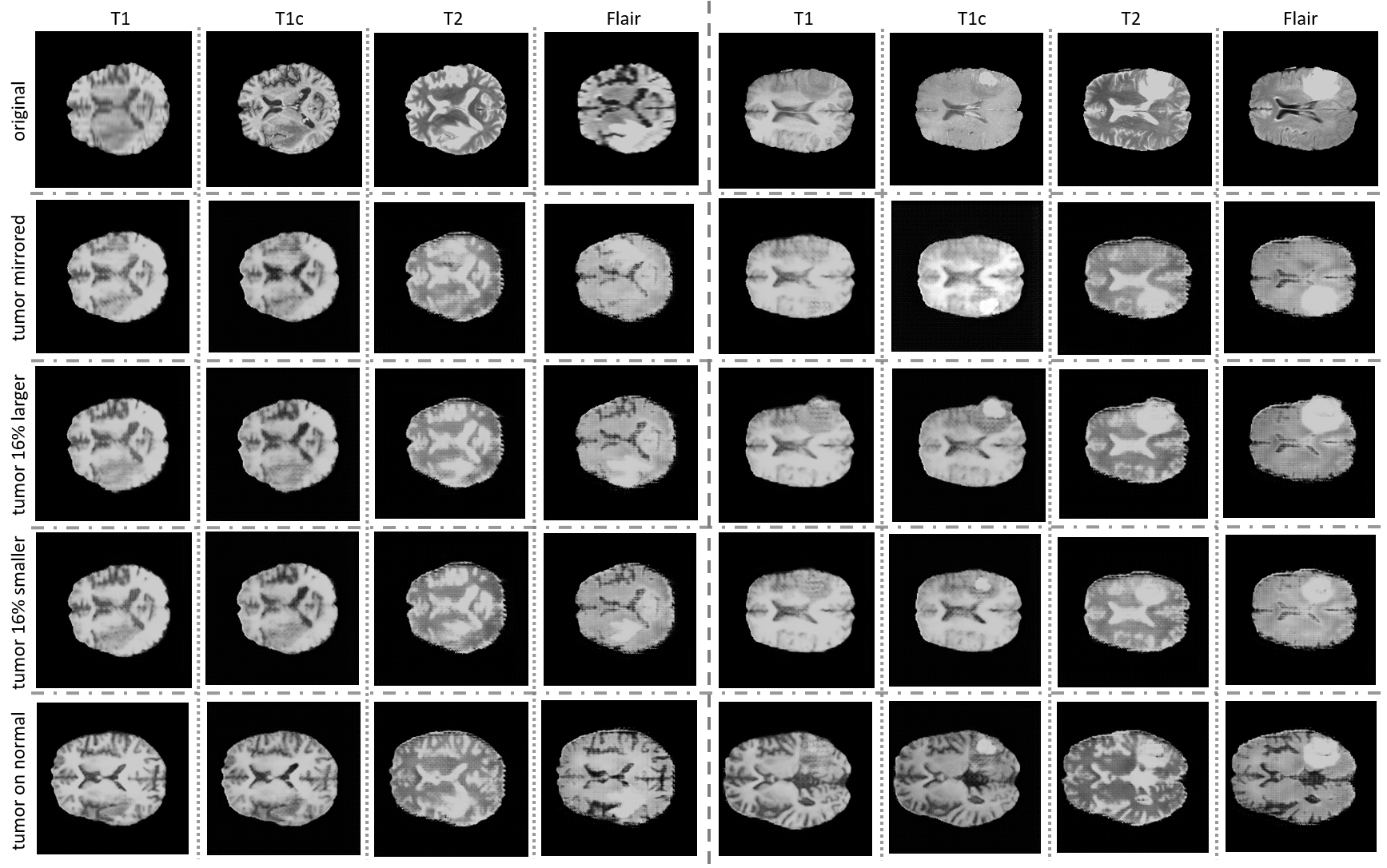}\\
  \includegraphics[width=1\linewidth]{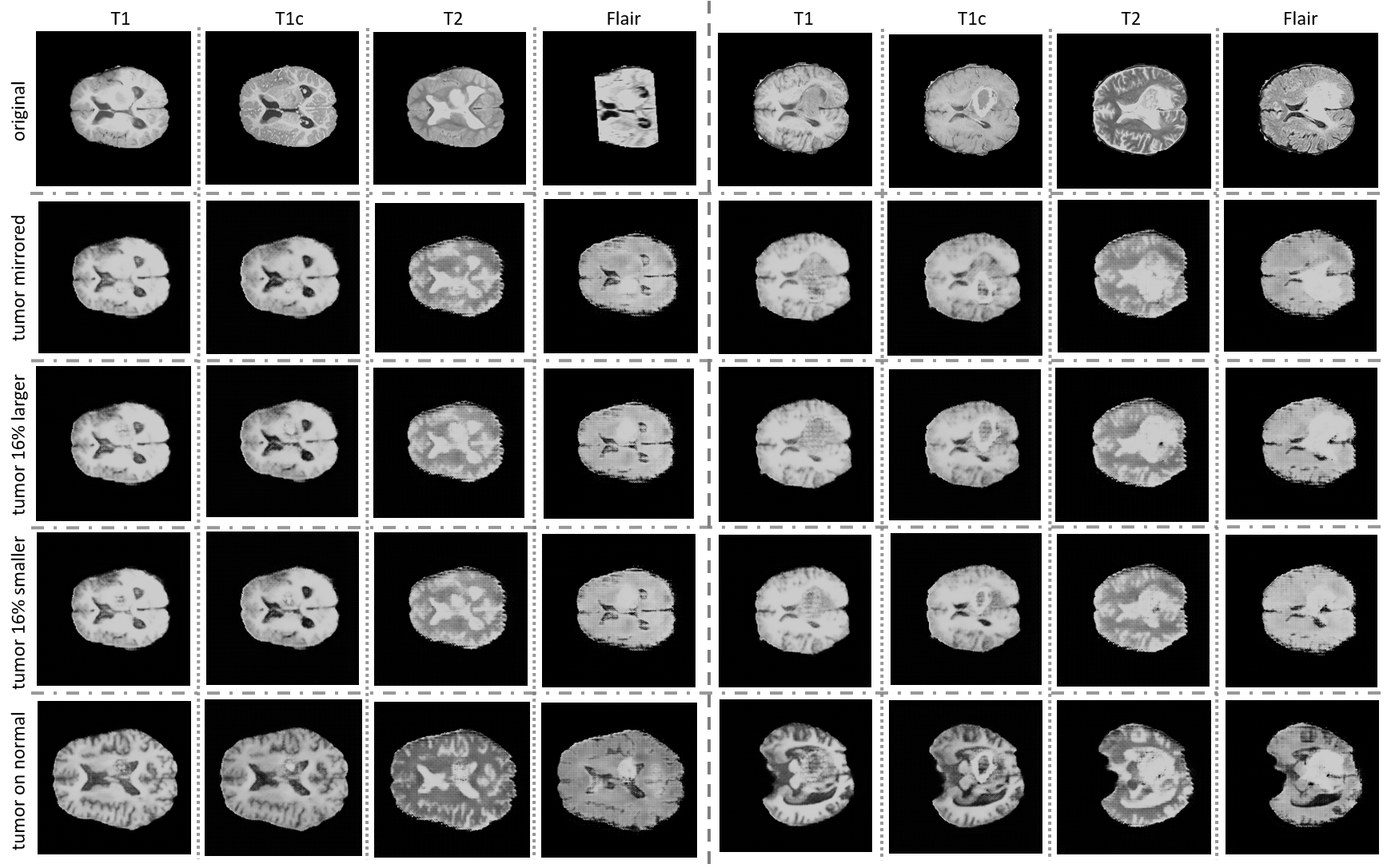}\\
  \caption{More examples of generated images.}
\label{fig:tm_manipulate_examples_2}
\vspace{-5pt}
\end{figure}

\end{document}